\documentclass[letterpaper, 10 pt, conference]{ieeeconf}  

\IEEEoverridecommandlockouts                              

\usepackage{enumerate}
\usepackage{amsmath}

\usepackage{amssymb}
\usepackage{graphicx} 
\usepackage{multirow}

\usepackage{caption}
\usepackage{bbding} 
\usepackage{stfloats}
\usepackage{cuted}
\usepackage{capt-of}
\usepackage{algorithmicx,algorithm}
\usepackage{color}
\usepackage{colortbl}
\usepackage{xcolor}
\usepackage{subfigure}
\usepackage{wrapfig}
\usepackage{listings}
\usepackage{cite}

\usepackage{listings}
\usepackage{framed}
\definecolor{lightblue}{rgb}{0.68, 0.85, 0.9} 
\definecolor{lightyellow}{rgb}{1, 1, 0.6} 

\usepackage[pagebackref,breaklinks,colorlinks]{hyperref}

\title{\texttt{ManipVQA}:  Injecting Robotic Affordance and Physically Grounded Information into Multi-Modal Large Language Models}

\author{Siyuan Huang$^{*}$, Iaroslav Ponomarenko$^{*}$, Zhengkai Jiang, Xiaoqi Li, Xiaobin Hu \\ Peng Gao, Hongsheng Li, Hao Dong
\thanks{$^{*}$ indicates the equal contribution.}%
\thanks{Siyuan Huang, Peng Gao, and Hongsheng Li are with the Shanghai AI Laboratory. Iaroslav Ponomarenko, Xiaoqi Li and Hao Dong are with Peking University. Siyuan Huang is also with Shanghai Jiao Tong University and Hongsheng Li is also with the MMLab, CUHK. Zhengkai Jiang is with UCAS and Xiaobin Hu is with TUM.}%
}

\begin{document}

\maketitle
\thispagestyle{empty}
\pagestyle{empty}

\begin{abstract}
While the integration of Multi-modal Large Language Models (MLLMs) with robotic systems has significantly improved robots' ability to understand and execute natural language instructions, their performance in manipulation tasks remains limited due to a lack of robotics-specific knowledge. Conventional MLLMs are typically trained on generic image-text pairs, leaving them deficient in understanding affordances and physical concepts crucial for manipulation. To address this gap, we propose \texttt{ManipVQA}, a novel framework that infuses MLLMs with manipulation-centric knowledge through a Visual Question-Answering (VQA) format. This approach encompasses tool detection, affordance recognition, and a broader understanding of physical concepts. We curated a diverse dataset of images depicting interactive objects, to challenge robotic understanding in tool detection, affordance prediction, and physical concept comprehension. To effectively integrate this robotics-specific knowledge with the inherent vision-reasoning capabilities of MLLMs, we leverage a unified VQA format and devise a fine-tuning strategy. This strategy preserves the original vision-reasoning abilities while incorporating the newly acquired robotic insights. Empirical evaluations conducted in robotic simulators and across various vision task benchmarks demonstrate the robust performance of \texttt{ManipVQA}. The code and dataset are publicly available at \href{https://github.com/SiyuanHuang95/ManipVQA}{https://github.com/SiyuanHuang95/ManipVQA}.
\end{abstract}
    
\section{\textbf{Introduction}}
\label{sec:intro}

\begin{figure}[th]
  \centering
   \vspace{0.1cm}
   \includegraphics[width=1\linewidth]{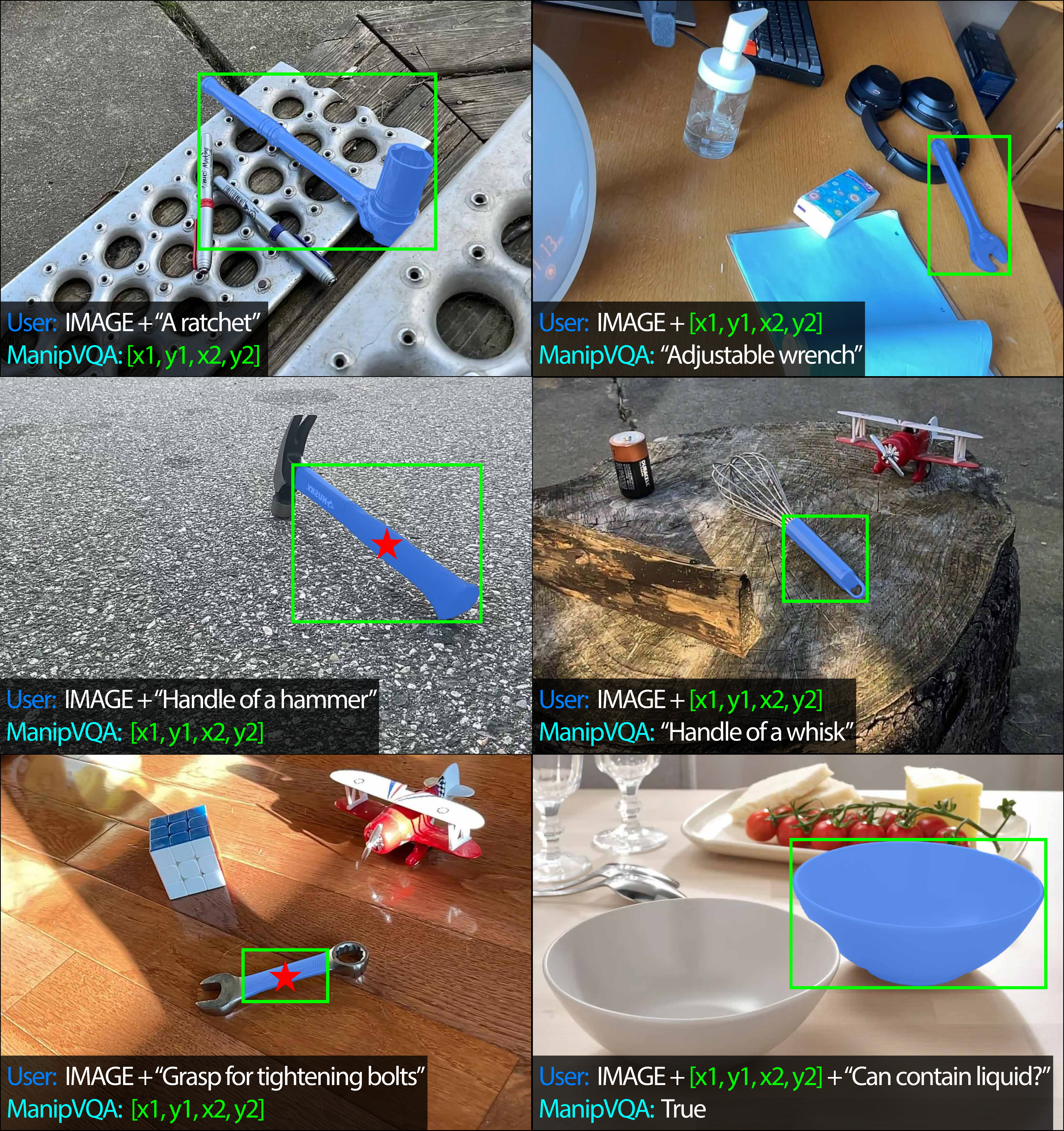}
   \caption{The \texttt{ManipVQA} model generates predictions in a unified VQA format, combining tool-object and affordance detection, affordance grounding, and an understanding of physical object properties. The displayed masks are created using SAM-HQ, based on the bounding boxes predicted by \texttt{ManipVQA}. The contact point, represented by a red star, indicates the geometric center of the bounding box.}
   \label{fig:the_ability_in_vqa_format}
   \vspace{-0.7cm}
\end{figure}
Recently, Multi-modal Large Language Models (MLLMs) such as OpenAI's GPT-4V~\cite{achiam2023gpt}, Google’s Gemini and SPHINX-X~\cite{gao2024sphinx}, have significantly advanced the capabilities for understanding and interpreting both text and images. These models achieve this improvement by aligning multi-modal encoders with Large Language Models (LLMs) through training on numerous text-image pairs or interleaved text-image pairs to enhance a comprehensive understanding of both modalities. As a result, MLLMs exhibit promising potential in addressing common sense reasoning and demonstrate remarkable generalization in vision tasks. However, the application of these models in manipulation tasks~\cite{huang2023voxposer,li2023manipllm}, such as robotic affordance understanding and the design of associated robotic systems, continues to present significant challenges.

Robotic manipulation~\cite{jia2024towards,li2023manipllm} is defined as a robot's capacity to perceive its environment and discern the potential actions applicable to various objects. The direct application of existing MLLMs, pre-trained on common scene reasoning, to robotic manipulation, does not yield satisfactory performance due to the absence of low-level action samples in their pretraining data. Previous research has addressed robotic affordance by prompting MLLMs to process scene images and then generate a sequence of robotic actions, but the performance has been suboptimal. Affordance grounding~\cite{qian2024affordancellm} aims to localize the regions in objects where actions are possible. This task also faces the challenge of establishing an explicit link with object parts due to the diversity of interactive affordances. Prior research~\cite{xia2023kinematic} employed LLMs to get the answer on affordance with prompt engineering. Despite the potential utility of MLLMs in robotic applications, their effectiveness is limited by several challenges. Classic MLLMs~\cite{gao2024sphinx} are generally trained on generic image-text pairs, thereby lacking critical robotic knowledge for understanding object affordances and their physical properties. This deficiency in specialized knowledge impedes their performance in manipulation tasks, consequently restricting the types of tasks that robots can execute and the precision of their execution. The gap between the capabilities of MLLMs and the demands of robotic systems constitutes a significant challenge that needs to be addressed.

To address this gap, we present \texttt{ManipVQA}, an innovative framework devised to equip MLLMs with manipulation-centric knowledge via a Visual Question-Answering paradigm. This integration is achieved through a unified VQA approach and a fine-tuning strategy that preserves the original vision-reasoning capabilities of MLLMs while infusing them with critical insights aimed at robotic tasks, enhancing tool detection, affordance recognition, and physical concepts understanding. To achieve this, we collect a diverse set of images featuring interactive objects, thus encompassing a broad range of challenges related to object detection, affordance, and physical concept prediction.

Empirical assessments performed in robotic simulators and across various vision task benchmarks substantiate the robust performance of \texttt{ManipVQA}. Our research makes several significant contributions to the fields of robotics and machine learning. First, we propose a novel approach to robotic manipulation and affordance understanding tasks, which addresses the shortcomings of existing methodologies. Second, we are dedicated to fostering the research community and have made our datasets, codes, and models publicly available. 

\vspace{-0.05cm}
\section{\textbf{Related Work}}

\begin{figure*}[t]
  \centering
   \vspace{-0.5cm}
  
   \includegraphics[width=1\linewidth]{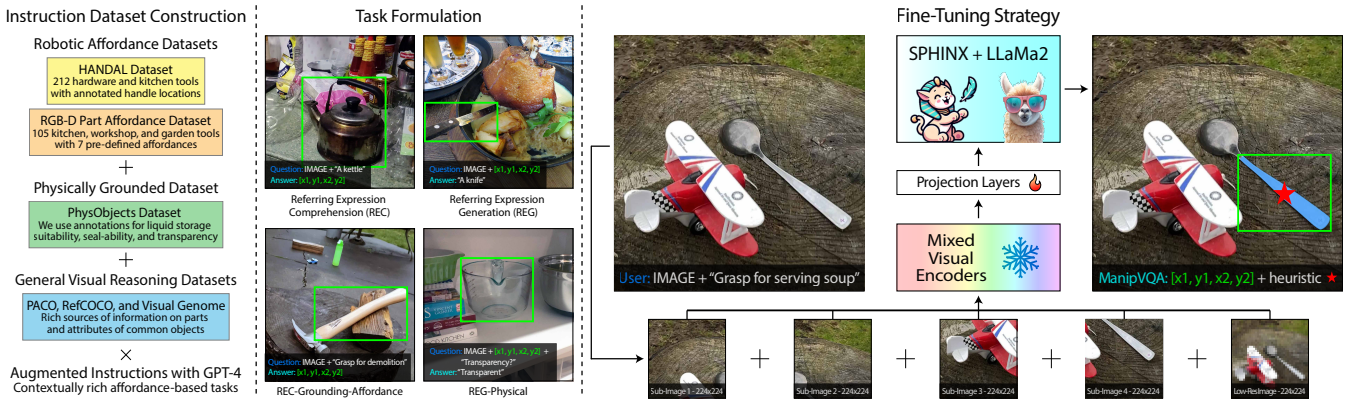}
   \caption{Overview of \texttt{ManipVQA}: We created a comprehensive vision-language dataset by merging existing datasets and expanding affordance grounding tasks with ChatGPT. To ensure consistency with existing VQA datasets, we adopted a similar VQA format. Using this dataset, we then fine-tuned an MLLM. When combined with a heuristic policy, the enhanced MLLM can perform a wide range of tasks, including complex manipulation tasks.}
   \label{fig:method}
   \vspace{-0.5cm}
\end{figure*}

\subsection{\textbf{Large-Object Datasets}}
The field of computer vision has been significantly advanced by datasets such as ImageNet~\cite{5206848} and PACO~\cite{ramanathan2023paco}, which have been crucial for progress in image classification, object detection, and segmentation. While these datasets enhance object instance detection and segmentation and provide a broad semantic understanding of objects, they often overlook the granularity of object parts and attributes, leading to a gap in robotic applications. To address this gap, datasets like HANDAL~\cite{guo2023handal} and PhysObjects~\cite{gao2023physically} have been introduced, focusing on robotic needs by providing annotations for part masks, object attributes, and affordances. HANDAL emphasizes manipulable object pose estimation with 308k annotated frames, while PhysObjects offers detailed annotations of household object properties. To unify these advancements, our \texttt{ManipVQA} merges these datasets into a cohesive VQA dataset format, granting models the ability to perform visual reasoning on general images as well as affordance reasoning for manipulation tasks, thereby enhancing the model's visual understanding for robotic applications.

\subsection{\textbf{Multi-Modal Large Language Model}}
Building on the foundation of extensive LLMs like LLaMa~\cite{touvron2023llama}, which exhibit remarkable language reasoning capabilities across various tasks, MLLMs have significantly expanded the scope of language processing by incorporating the ability to understand visual stimuli~\cite{lin2023sphinx,gao2024sphinx}. A notable innovation in this field is SPHINX~\cite{lin2023sphinx}, which excels in multi-modal tasks by aligning ensembled visual features with language embeddings through projection layers. Despite these advances, the application of MLLMs to robotic manipulation remains in its early stages. Some recent initiatives such as~\cite{li2023manipllm, li2023vision, qian2024affordancellm} have attempted to integrate robotic-domain knowledge into MLLMs. However, these efforts often overlook critical physical information~\cite{li2023manipllm} or do not fully embody a robotic-centric approach~\cite{qian2024affordancellm}. In contrast, our proposed \texttt{ManipVQA} integrates essential physical knowledge and affordance reasoning capabilities from a robotic perspective into MLLMs. We present this information in a direct, human-readable, purely linguistic format, enhancing the utility of MLLMs for practical robotic applications.

\subsection{\textbf{Language-Driven Robotics Manipulation}}
Instruction-based policies in robotics serve as a bridge between high-level human communication and low-level robotic control, offering intuitive interaction and the potential for skill transfer and complex task planning. One approach \cite{shridhar2022cliport, shridhar2023perceiver} involves using a pre-trained text encoder to extract text embeddings that encapsulate linguistic knowledge, which are then used to inform robotic policies. For example, CLIPort has obtained the ability to understand semantics and manipulate objects by encoding text input through CLIP\cite{radford2021learning}. Another research direction leverages the reasoning capabilities of LLMs for planning~\cite{wu2023tidybot, cai2023bridging} or even directly generating the policy codes~\cite{huang2023voxposer,huang2023instruct2act, ma2023eureka} by prompt engineering. For example, CaP\cite{liang2022code} generates policy codes with detailed comments and context-specific examples to guide LLM output. Furthermore, some studies have trained large-scale policy models~\cite{brohan2023can,driess2023palm, brohan2023rt} using extensive robotic datasets, aiming to learn robotic control policies from diverse data directly. Despite the progress in these areas, a gap remains in the explicit consideration of affordance grounding and physical reasoning, which are essential for proficient manipulation in robotics. Aiming to bridge this gap, our \texttt{ManipVQA} injects robotic affordance and physically grounded information into MLLM while preserving its original reasoning ability.

\vspace{-0.075cm}
\section{\textbf{Method}}

\label{sec:method}

\subsection{\textbf{Modeling of Affordances and Physical Concepts}}
\label{sec:modeling_affordances}
In robotic manipulation, understanding and modeling object affordances is essential for enabling robots to interact effectively with their environment. We extend upon the graspable affordance model $\mathbf{A_{grasp}}$ for common tool objects $\mathbf{O_{tool}}$, forming the tuple $(\mathbf{A_{grasp}}, \mathbf{O_{tool}})$, as delineated by HANDAL~\cite{guo2023handal}. In our representation, each (partial) object is paired with a bounding box, described by the coordinates $[x_{min}, y_{min}, x_{max}, y_{max}]$, which specify the top-left and bottom-right corners, respectively.

To ensure our model's adaptability across different contexts, we normalize these coordinates relative to the image dimensions, enhancing the model's generalizability and robustness. Furthermore, we recognize that affordances may vary depending on the specific task $\mathbf{T}$ at hand. For instance, distinct regions of a tool may be utilized for different functions. To address this, we form a tuple $(\mathbf{A_{T}}, \mathbf{O_{tool}})$, which associates task-specific affordances $\mathbf{A_{T}}$ with the tool object. The task $\mathbf{T}$ is succinctly described using a brief natural language sentence to encapsulate its functionality.

In addition to affordances, we incorporate the modeling of physical concepts, denoted as $\mathbf{P}_{i}$. These concepts are quantified using discrete levels or boolean values, drawing inspiration from the methodology presented in PhysObjects~\cite{gao2023physically}. Each physical concept $\mathbf{P}_{i}$ is linked to its corresponding object $\mathbf{O}$, resulting in the tuple $(\mathbf{P}_{i}, \mathbf{O})$. Table~\ref{tab:list_phy_concept} enumerates the physical concepts under consideration, alongside their respective descriptions. These concepts include but are not limited to, transparency, liquid storage capacity, and seal-ability. Each concept is crucial for the robot's ability to perceive and interact with various objects within its environment.


\begin{table}[]
\centering
\resizebox{0.48\textwidth}{!}{%
\begin{tabular}{c|c|c} \hline

\textbf{Phys. Concepts} &
\textbf{Descriptions} &
\textbf{Type} \\ \hline \hline

\multicolumn{1}{l|}{Transparency} & 
\multicolumn{1}{l|}{Object's light transmission ability} &
Levels \\ 

\multicolumn{1}{l|}{Liquids Storage} &
\multicolumn{1}{l|}{Object's liquid-holding capacity} &
Bool \\ 

\multicolumn{1}{l|}{Seal-ability} & 
\multicolumn{1}{l|}{Object's closure state} &
Bool \\ \hline \hline

\end{tabular}%
}
\caption{Physical concepts used in \texttt{ManipVQA}.}
\label{tab:list_phy_concept}
\vspace{-0.5cm}
\end{table}

\subsection{\textbf{Instruction Dataset Construction}}
In the development of the \texttt{ManipVQA} model, we focused on creating a comprehensive training dataset by combining various publicly available datasets. The goal was to equip the model with a strong understanding of robotic affordances, advanced visual reasoning, and a solid foundation in physically grounded knowledge.

\subsubsection{\textbf{Robotic Affordance Datasets}}
The HANDAL Dataset~\cite{guo2023handal} serves as a foundation base for imparting the \texttt{ManipVQA} model with the ability to discern objects and identify graspable components. It comprises over 212 real-world objects, each with annotated handle locations. To further refine the model's capability in understanding complex affordances, we integrated the RGB-D Part Affordance Dataset~\cite{Myers:ICRA15}, which differentiates seven pre-defined affordances such as grasp, cut, scoop, contain, pound, support, and wrap-grasp. While the AGD20K Dataset~\cite{luo2022learning} represents another extensive affordance resource, it is predominantly labeled with broad human-centric action terms, such as \textit{drink} or \textit{sit-on}. Therefore, we have excluded AGD20K from the training regime and instead selected it for zero-shot evaluation, assessing the model's generalization capabilities.

\subsubsection{\textbf{Physically Grounded Dataset}}
The PhysObjects Dataset~\cite{gao2023physically} was employed to infuse the model with a nuanced understanding of physical properties. This dataset features eight core physical concepts, but we selectively utilized annotations relevant to our focus—specifically, those related to liquid storage suitability, seal-ability, and transparency as listed in Table~\ref {tab:list_phy_concept}.

\subsubsection{\textbf{General Visual Reasoning Datasets}}
To maintain and extend the model’s capabilities in general visual reasoning, we incorporated datasets such as the PACO~\cite{ramanathan2023paco}, RefCOCO~\cite{kazemzadeh2014referitgame}, and the Visual Genome~\cite{krishna2016visual}, which provide a rich source of information on parts and attributes of common objects.

\subsubsection{\textbf{Augmented Instructions with GPT-4}}
Considering the limitations of existing annotations in the robotic affordance datasets—primarily confined to partial masks in~\cite{guo2023handal} and basic action types with associated masks in~\cite{Myers:ICRA15}, we employed GPT-4 to generate complex and contextually rich affordance-based tasks. This augmentation enables our model to not only learn explicit commands but also to interpret and execute complex, implicitly defined tasks. Listing~\ref{lst:gpt_prompt_for_grounding} showcases examples of the prompts used for generating these advanced affordance grounding tasks.


\begin{lstlisting}[
float=t,
language=bash,
floatplacement=htbp,
frame=TRBL,
frameround=tftf,
belowskip=-2\baselineskip,
basicstyle=\ttfamily\scriptsize,
breakatwhitespace=true,
breaklines=true,
captionpos=b,
columns=flexible,
keepspaces=true,
tabsize=2,
showspaces=false,
showstringspaces=false,
showtabs=false,
label={lst:gpt_prompt_for_grounding},
caption= Example prompt for generating an affordance grounding tasks for a specific tool object using ChatGPT.,
abovecaptionskip=0pt,
belowcaptionskip=0pt]
<@\textbf{Role:}@>
<@ You are an analytical assistant specializing in robotic @>
<@ affordance grounding. Your expertise is in creating @>
<@ tasks that facilitate the training of robotic policies, @> 
<@ enabling robots to reason about task execution, such as @> 
<@ determining the appropriate part of an object to grasp. @> 

<@\textbf{Task Description:}@>
<@ You will be provided with the name of a tool that can @> 
<@ be attached to a robotic arm. The robot is expected to @>
<@ use this tool to perform a variety of everyday tasks. @> 
<@ Along with the tool name, you will receive a list of @>  
<@ tasks that have already been generated for this tool. @>

<@\textbf{Guidelines:}@>
<@ \textbf{Diversity}: Aim for a wide range of tasks, ensuring that @>
<@ there is no overlap with previous ones. @>
<@ \textbf{Daily Tasks}: Tasks should be common and representative @>
<@ of the ones encountered in daily life. @>
<@ \textbf{Leakage Avoidance}: Ensure that the generated tasks do @> 
<@ not explicitly mention the name of the tool object. @>

<@\textbf{Examples:}@>
<@~~\textcolor{teal}{...}@>

<@\textbf{Instruction:} @>
<@ With the provided \textcolor{blue}{OBJECT\_NAME}, generate five new @>
<@ affordance grounding tasks. Use the \textcolor{blue}{HISTORY} of @>  
<@ generated tasks as a reference to ensure compliance @> 
<@ with the diversity guideline. Output should be in the @> 
<@ JSON format with the object name as the key. @>
\end{lstlisting}

\begin{table*}[]
\vspace{-0.5cm}
\centering
\resizebox{\textwidth}{!}{%
\begin{tabular}{c|c|l|c|c}
\hline

\textbf{Capabilities} &
\textbf{Tasks} &
\multicolumn{1}{c|}{\textbf{Examples of Task Templates}} &
\textbf{Source} &
\textbf{Num.} \\ \hline \hline

\multicolumn{1}{l|}{Gen. Visual Reasoning} &
\multicolumn{1}{l|}{REC/REG: Object} &
\begin{tabular}[l]{@{}l@{}}
    \textcolor{blue}{\texttt{User}}: Please provide a short description of this region: \textbf{BBox}. \\ \textcolor{teal}{\texttt{ManipVQA}}: A ratchet.
\end{tabular} &

H/P/C/V &
36K \\ \hline \hline

Robotic &
\multicolumn{1}{l|}{REC/REG: Affordance} &
\begin{tabular}[c]{@{}l@{}}
    \textcolor{blue}{\texttt{User}}: Please provide bounding box coordinates of this region: handle of a screwdriver. \\ 
    \textcolor{teal}{\texttt{ManipVQA}}: \textbf{BBox}.
\end{tabular} &  

H/R &
26K \\ \cline{2-5} 

\multirow{-2}{*}{Aff. Understanding} &
\multicolumn{1}{l|}{REC-Grounding: Affordance} &
\begin{tabular}[c]{@{}l@{}}
    \textcolor{blue}{\texttt{User}}: Please provide bounding box coordinates of this region: grasp for tightening bolts. \\   \textcolor{teal}{\texttt{ManipVQA}}: \textbf{BBox}.
\end{tabular} &
H/R &
15K \\ \hline \hline

\multicolumn{1}{l|}{Phys. Gr. Understanding} &
\multicolumn{1}{l|}{REG-Phys: Liq./Seal./Transp.} &
\begin{tabular}[c]{@{}l@{}}
    \textcolor{blue}{\texttt{User}}: Please provide a short description of whether this object can contain liquid: \textbf{BBox}. \\ \textcolor{teal}{\texttt{ManipVQA}}: True. \end{tabular} &
Phys &
7K \\ \hline \hline

\end{tabular}%
}
\caption{Overview of the \texttt{ManipVQA} Dataset: This table summarizes the tasks, their associated capabilities, example templates, and the number of samples in each dataset. The capabilities are represented by the abbreviations: ``Gen." for \textit{General}, ``Aff." for \textit{Affordance}, and ``Phys. Gr." for \textit{Physically Grounded}. The source datasets are identified by the acronyms: H for HANDAL~\cite{guo2023handal}, P for PACO~\cite{ramanathan2023paco}, C for RefCOCO~\cite{kazemzadeh2014referitgame}, R for RGB-D Part Affordance Dataset~\cite{Myers:ICRA15}, and Phys for PhysObjects~\cite{gao2023physically}. The bounding box (BBox) format is represented by $[x_{min}, y_{min}, x_{max}, y_{max}]$.}

\label{tab:Task_Templates_Examples}
\vspace{-0.5cm}
\end{table*}

\subsection{\textbf{Task Formulation}}
\label{sec:task_formulation}
The \texttt{ManipVQA} training protocol integrates a pair of principal vision-language tasks: Referring Expression Comprehension (REC) and Referring Expression Generation (REG). Following~\cite{chen2023shikra}, REC involves the model receiving an image accompanied by a natural language description and subsequently predicting the bounding box coordinates that delineate the specified target within the image. Conversely, REG prompts the model to produce a descriptive natural language statement about an area within an image, defined by provided bounding box coordinates. To advance \texttt{ManipVQA}'s proficiency in recognizing robotic affordances and discerning object physical properties, we have augmented the task framework with:

\begin{itemize}
    \item \textit{\textbf{REC-Grounding-Affordance}}: This task refines the model's capacity to identify functional parts of objects based on their usage descriptions. It presents the challenge of localizing these parts without directly naming the object or its components, a step towards intuitive affordance recognition for robots.
    \item \textit{\textbf{REC-Physical}}: This task broadens the model's attribute recognition by requiring it to pinpoint objects based on their physical properties and to engage with related inquiries. This is essential for detailed robotic perception and manipulation.
\end{itemize}

These additional tasks enhance the core REC and REG tasks, together cultivating a robust skill set tailored for practical robotic deployment. Detailed instances of these task formats can be found in Table~\ref{tab:Task_Templates_Examples}.

\subsection{\textbf{MLLM Fine-Tuning Strategy}}
\subsubsection{\textbf{Model Architecture}} 
We adopt the MLLM, SPHINX~\cite{lin2023sphinx} as our primary architecture, with LLaMA2~\cite{touvron2023llama} serving as the language backbone. Given the necessity for both global and local visual grounding in robotic tasks, we integrate the visual encoder from CLIP~\cite{radford2021learning} to extract local semantic features and the Q-Former~\cite{li2023blip} for summarizing visual features. Spatial alignment is facilitated using projection layers, and global features are merged with local ones through channel-wise concatenation. We acknowledge that the standard image resolution for pre-trained visual encoders, typically $224 \times 224$, is insufficient for detailed visual perception. This limitation is particularly significant for robotic affordance reasoning, which often requires fine-grained visual grounding of object parts, such as tool handles or machine buttons. To augment \texttt{ManipVQA}'s region-level grounding capabilities, we employ a sub-images patching strategy following ~\cite{lin2023sphinx}. Specifically, we partition a $448 \times 448$ image into four $224 \times 224$ sub-images taken from each corner, thereby preserving intricate visual details. The resulting image tokens are then positioned before the language instructions to provide visual context for the ensuing prompts.

\subsubsection{\textbf{Fine-Tuning Strategy}} 
As elucidated in Sec.~\ref{sec:modeling_affordances}, we model both affordance and physical concepts within natural language representations and training samples are formatted in line with the general VQA framework, as delineated in Sec.~\ref{sec:task_formulation}. As a result, the training objective employs a unified cross-entropy loss, diverging from the approaches in ~\cite{qian2024affordancellm, li2023manipllm}. To maintain the model's broad visual reasoning proficiency, we amalgamate general visual reasoning exercises with tasks specific to robotics. 

\section{\textbf{Experiments}}
\label{sec:experiments}
\subsection{\textbf{Implementation Details}}
\label{sec:implementation_details}
\subsubsection{\textbf{Training Details}}
We fine-tuned the \texttt{ManipVQA} model using the SPHINX framework~\cite{lin2023sphinx} on eight NVIDIA A100 (80GB) GPUs.  The fine-tuning was completed in a single epoch, which took approximately 6 hours. During this phase, the visual encoders were kept frozen to maintain the integrity of the pre-trained features. The pre-trained model was the SPHINX-1K, obtained from the official repository. Training was conducted with a batch size of 4 and a learning rate set to $2 \times 10^{-5}$.

\begin{table*}[!htbp]
\vspace{-0.675cm}
\centering
\resizebox{\textwidth}{!}{%
\begin{tabular}{l|c|ccccccccc|ccccccccccc}
  
  \hline 
  & & 
  \multicolumn{9}{c|}{\textbf{Hardware Tools}} &
  \multicolumn{8}{c|}{\textbf{Kitchen Tools}} &
  \multirow{2}{*}{\textbf{AVG}} \\ \cline{3-19} 
  
  \multirow{-2}{*}{\textbf{Method}} & \multirow{-2}{*}{\textbf{Task}}   &
  \multicolumn{1}{c|}{\textbf{Ha}} &
  \multicolumn{1}{c|}{\textbf{Pf}} &
  \multicolumn{1}{c|}{\textbf{Ps}} &
  \multicolumn{1}{c|}{\textbf{Pl}} &
  \multicolumn{1}{c|}{\textbf{Pd}} &
  \multicolumn{1}{c|}{\textbf{Ra}} &
  \multicolumn{1}{c|}{\textbf{Sd}} &
  \multicolumn{1}{c|}{\textbf{Wa}} &
  \multicolumn{1}{c|}{\textbf{W.c}} &
  \multicolumn{1}{c|}{\textbf{La}} &
  \multicolumn{1}{c|}{\textbf{Mc}} &
  \multicolumn{1}{c|}{\textbf{Mug}} &
  \multicolumn{1}{c|}{\textbf{Pan}} &
  \multicolumn{1}{c|}{\textbf{Sp}} &
  \multicolumn{1}{c|}{\textbf{St}} &
  \multicolumn{1}{c|}{\textbf{Ut}} &
  \multicolumn{1}{c|}{\textbf{Wh}} &
  \multicolumn{1}{c}{} \\ \hline \hline

HANDAL &  & 
  0.76 &
  0.48 &
  0.42 &
  0.77 &
  0.74 &
  0.67 &
  0.60 &
  0.63 &
  0.79 &
  0.81 &
  0.53 &
  0.53 &
  0.80 &
  0.46 &
  0.66 &
  0.74 &
  0.79 &
  \cellcolor[HTML]{C0C0C0} 0.66 \\ 

\textbf{Ours} &  \multirow{-2}{*}{Obj-B} &
  0.96 &
  0.96 &
  0.94 &
  0.89 &
  0.98 &
  0.93 &
  0.94 &
  0.93 &
  0.92 &
  0.97 &
  0.95 &
  0.90 &
  0.97 &
  0.90 &
  0.94 &
  0.88 &
  0.95 &
  \cellcolor[HTML]{C0C0C0} \textbf{0.94} \\ \hline 

\textbf{Ours} & Aff-B &
  0.70 &
  0.67 &
  0.64 &
  0.71 &
  0.33 &
  0.71 &
  0.76 &
  0.56 &
  0.69 &
  0.82 &
  0.51 &
  0.52 &
  0.59 &
  0.54 &
  0.81 &
  0.72 &
  0.65 &
  \cellcolor[HTML]{C0C0C0} 0.64 \\ \hline \hline

HANDAL &  & 
  0.62 &
  0.38 &
  0.27 &
  0.67 &
  0.75 &
  0.44 &
  0.56 &
  0.48 &
  0.52 &
  0.67 &
  0.49 &
  0.55 &
  0.81 &
  0.40 &
  0.62 &
  0.57 &
  0.75 &
  \cellcolor[HTML]{C0C0C0} 0.56 \\ 

LISA &  & 
  0.84 &
  0.63 &
  0.66 &
  0.70 &
  0.93 &
  0.68 &
  0.76 &
  0.79 &
  0.80 &
  0.77 &
  0.67 &
  0.78 &
  0.83 &
  0.77 &
  0.58 &
  0.79 &
  0.84 &
  \cellcolor[HTML]{C0C0C0} \textbf{0.75} \\ 

\textbf{Ours} & \multirow{-3}{*}{Obj-M}& 
  0.71 &
  0.70 &
  0.61 &
  0.65 &
  0.57 &
  0.52 &
  0.82 &
  0.66 &
  0.62 &
  0.57 &
  0.53 &
  0.55 &
  0.43 &
  0.55 &
  0.46 &
  0.58 &
  0.55 &
  \cellcolor[HTML]{C0C0C0} 0.58 \\ \hline \hline

LISA &   & 
  0.67 &
  0.59 &
  0.56 &
  0.54 &
  0.43 &
  0.48 &
  0.62 &
  0.62 &
  0.61 &
  0.41 &
  0.41 &
  0.56 &
  0.45 &
  0.58 &
  0.40 &
  0.49 &
  0.49 &
  \cellcolor[HTML]{C0C0C0} 0.59 \\ 

\textbf{Ours} & \multirow{-2}{*}{Aff-M}  & 
  0.75 &
  0.70 &
  0.54 &
  0.53 &
  0.44 &
  0.56 &
  0.80 &
  0.65 &
  0.68 &
  0.62 &
  0.60 &
  0.52 &
  0.62 &
  0.65 &
  0.64 &
  0.58 &
  0.68 &
  \cellcolor[HTML]{C0C0C0} \textbf{0.62} \\ \hline \hline

LISA &    & 
  0.57 &
  0.42 &
  0.41 &
  0.48 &
  0.35 &
  0.41 &
  0.62 &
  0.54 &
  0.56 &
  0.37 &
  0.35 &
  0.43 &
  0.36 &
  0.39 &
  0.37 &
  0.41 &
  0.48 &
  \cellcolor[HTML]{C0C0C0} 0.44 \\ 

\textbf{Ours} & \multirow{-2}{*}{Gr-Aff-M}  & 
  0.82 &
  0.69 &
  0.49 &
  0.42 &
  0.42 &
  0.54 &
  0.83 &
  0.66 &
  0.64 &
  0.71 &
  0.66 &
  0.59 &
  0.73 &
  0.71 &
  0.78 &
  0.78 &
  0.63 &
  \cellcolor[HTML]{C0C0C0} \textbf{0.65} \\ \hline \hline
   
\end{tabular}%
}
\caption{Evaluation results on HANDAL Dataset. Task abbreviations are as follows: ``Obj." for \textit{Complete Object Detection}, ``Aff." for \textit{Robotic Affordance Detection}, and ``Gr" for \textit{Grounded Detection}. The letter ``B" denotes bounding box format, and ``M" denotes mask representation format. Object abbreviations are listed in sequence: Hammer, Pliers-Fixed Joint, Pliers-Slip Joint, Pliers-Locking, Power Drill, Ratchet, Screwdriver, Wrench-Adjustable, Wrench-Combinational, Label, Measuring Cup, Mug, Pan, Spatula, Strainer, Utensil, and Whisk.}
\label{tab:eval_handal_dataset}
\vspace{-0.5cm}
\end{table*}

\subsubsection{\textbf{Connected with Robotic Policy}} 
The objective of \texttt{ManipVQA} is to augment the generalizability of robotic control policies. It could be used in a language-free format during high-level decision-making for physically grounded knowledge. For affordance localization, while the initial bounding box identifies areas of potential manipulation, it may also encompass extraneous elements like background features. To achieve a more precise affordance map, we leverage the SAM-HQ variant~\cite{ke2024segment}, which uses the initial bounding box as a ``box prompt" for more accurate segmentation. Heuristic methods are then employed to determine the contact point. In our experiments, the ground truth (GT) surface normals, essential for rotation estimation, were presumed to be accessible either through RGB-D sensing or an alternative pre-trained model.

\subsection{\textbf{Experimental Setup}}
We aim to systematically evaluate the performance of our approach, encompassing both robotic affordance grounding and robotic manipulation tasks.

\subsubsection{\textbf{Robotic Affordance Detection}} 
Our evaluation is primarily conducted on the HANDAL Dataset~\cite{guo2023handal}, assessing both bounding box average precision (AP) and pixel-wise segmentation AP. Unlike the baseline model in the HANDAL, which detects only the whole object, our \texttt{ManipVQA} is capable of identifying both the entire object and its manipulable parts, namely the affordance regions. Additionally, we compare our approach to LISA~\cite{lai2023lisa}, which integrates LLM and SAM decoder.

\subsubsection{\textbf{Physical Concept Grounding}}
We utilize the PhysObjects Dataset~\cite{gao2023physically} to evaluate physical concept grounding capabilities. We benchmark \texttt{ManipVQA} against PG-InstructBLIP~\cite{gao2023physically}, a fine-tuned version of InstructBLIP on PhysObjects, and the latest and most advanced MLLM, GPT-4V. Due to the limited localization capacity of GPT-4V, we pair it with Set-of-Mark~\cite{yang2023set}, where bounding boxes and index numbers are explicitly annotated on the target objects within the input images.

\subsubsection{\textbf{General Affordance Grounding}}
Although \texttt{ManipVQA} is trained solely on a robotic affordance dataset, we are interested in its generalization capabilities on broader affordance grounding datasets, such as on AGD20K~\cite{luo2022learning}. This exploration is motivated by the robust reasoning and generalization potential of LLMs. Our method is evaluated on AGD20K and follows its metrics, including KLD, SIM, and NSS. We compared with AffordanceLLM~\cite{qian2024affordancellm}, Cross-View-AG~\cite{luo2022learning}, LOCATE~\cite{li2023locate} and 3DOI~\cite{qian2023understanding}.

\subsubsection{\textbf{Robotic Manipulation Tasks}}
We further integrate \texttt{ManipVQA} with a basic robotic control policy, as detailed in Section~\ref{sec:implementation_details}, and use the manipulation success rate as a metric to gauge its practicality in robotic manipulation tasks using the PartNet-Mobility Dataset within the SAPIEN~\cite{xiang2020sapien} simulator. Our experimental setup is modeled after ManipLLM~\cite{li2023manipllm}, utilizing identical metrics. However, we employ ground truth (GT) surface normals from the simulator for rotation estimation. We compare our method with Where2Act~\cite{mo2021where2act}, FlowBot3D~\cite{eisner2022flowbot3d}, and ManipLLM~\cite{li2023manipllm}. For additional details on the experimental settings, please refer to ManipLLM~\cite{li2023manipllm}.


\subsection{\textbf{Results}}
\subsubsection{\textbf{Robotic Affordance Detection Evaluation}}
As illustrated in Table~\ref{tab:eval_handal_dataset}, our \texttt{ManipVQA} achieves remarkable performance in the detection of both complete objects and their affordances in a unified framework. Enhanced by the SAM-HQ, our model attains superior results in tasks involving affordance detection and grounding with a mask. However, it falls short in complete object segmentation, which we attribute to a tendency of SAM-HQ to cause over-segmentation.

\begin{figure*}
\vspace{-0.5cm}
\centering
\includegraphics[width=1\linewidth]{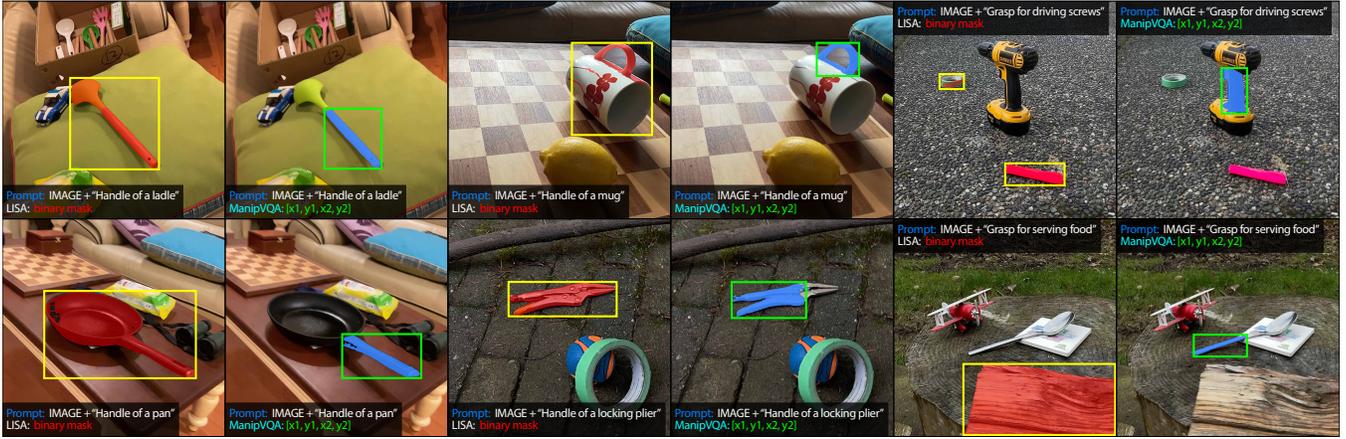}
\caption{Illustration of the affordance detection and grounding tasks using the HANDAL Dataset. The first four columns show the results for affordance detection, and the last two columns present the results for the affordance grounding task. In the affordance grounding task, a description of the tool's intended use is given instead of explicit part names. Blue masks are produced by SAM-HQ based on \texttt{ManipVQA}'s predicted bounding boxes, and red masks are LISA's output.}

\label{fig:affordance_on_handal}
\end{figure*}

\subsubsection{\textbf{Physical Concept Grounding Evaluation}}
Table~\ref{tab:eval_result_on_phy} presents the evaluation results on the PhysObjects~\cite{gao2023physically}. It is noteworthy that even the advanced MLLM GPT-4V encounters challenges with tasks that require an understanding of physical concepts. This struggle probably stems from its limited capacity for precise localization and a deficiency in visual physical reasoning. With the integration of SoM~\cite{yang2023set}, the GPT-4V has better localization ability while its performance remains suboptimal. Our \texttt{ManipVQA} outperforms PG-InstructCLIP~\cite{gao2023physically} which is also fine-tuned on  PhysObjects, and we hypothesize that this enhanced performance can be attributed to the more powerful LLM and the ensemble of vision encoders deployed within our MLLM.

\begin{table}[h]
\centering
\resizebox{0.48\textwidth}{!}{%
\begin{tabular}{c|c|c|c|c} \hline

\textbf{Methods} &
\textbf{Trans.} &
\textbf{Liquid Stor.} &
\textbf{Seal.} &
\textbf{AVG} \\ \hline \hline

\multicolumn{1}{l|}{GPT-4V} & 
35.0 &
52.8 &
49.7 &
45.8 \\ 

\multicolumn{1}{l|}{SoM~\cite{yang2023set} + GPT-4V} & 
34.5 &
55.5 &
53.1 &
47.7 \\ 

\multicolumn{1}{l|}{PG-InstructBLIP~\cite{gao2023physically}} & 
83.8 &
\textbf{89.1} &
80.6 &
84.5 \\  \hline

\multicolumn{1}{l|}{\textbf{Ours}} & 
\textbf{93.5} &
85.6 &
\textbf{91.7} &
\textbf{90.3} \\ \hline \hline

\end{tabular}%
}
\caption{Physcial evaluation results on PhysObject Dataset~\cite{gao2023physically}.}
\vspace{-0.5cm}
\label{tab:eval_result_on_phy}
\end{table}

\subsubsection{\textbf{General Affordance Grounding}} 
\begin{figure}
\vspace{-0.1cm}
\centering
\includegraphics[width=1\linewidth]{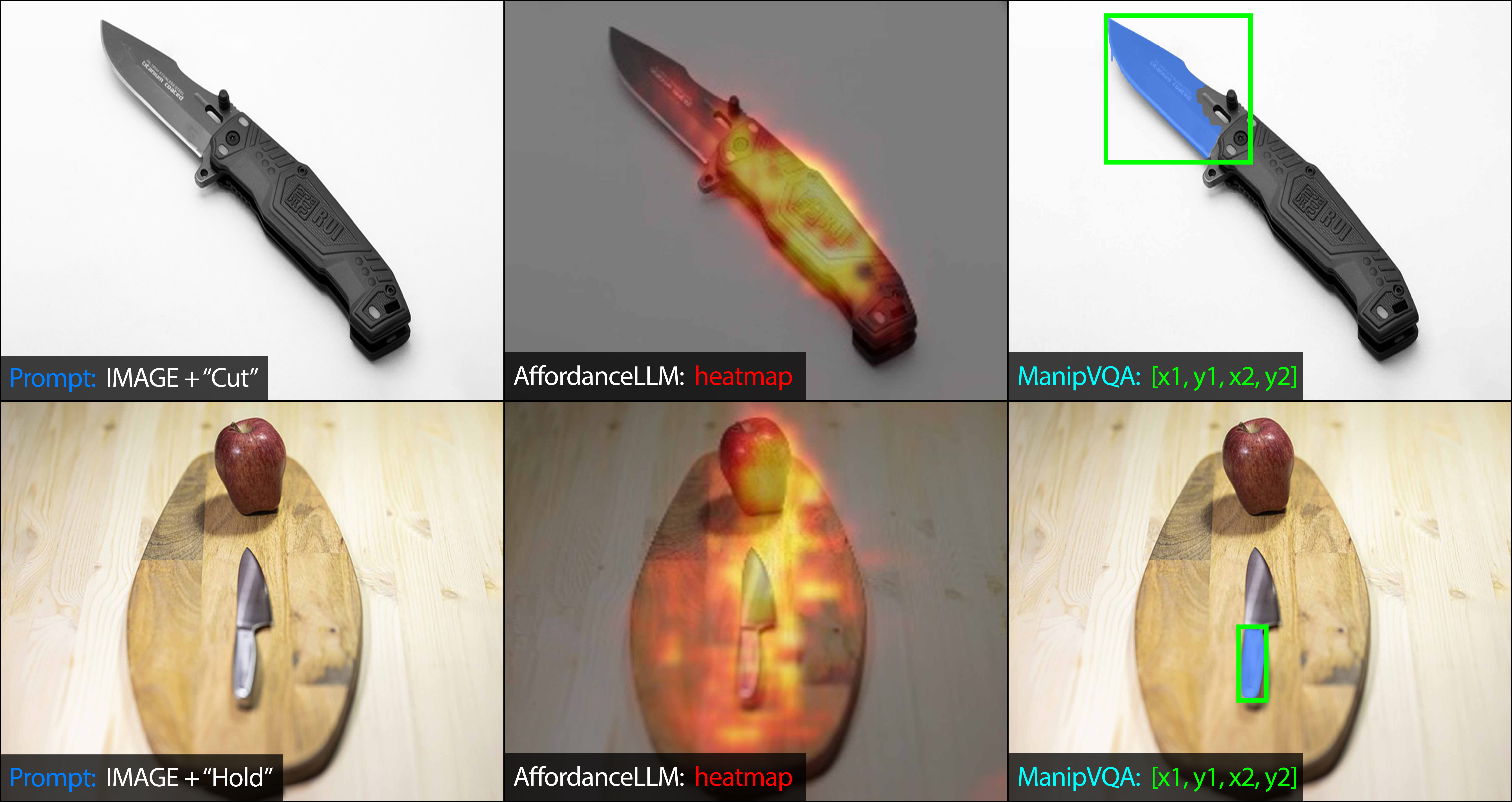}
\caption{Strong reasoning capability of \texttt{ManipVQA} to distinguish subtle features within affordances of the same object class. Blue masks are produced by SAM-HQ based on \texttt{ManipVQA}'s predicted bounding boxes.}
\vspace{-0.6cm}
\label{fig:affordance_knife_with_different_instr}
\end{figure}

Table~\ref{tab:general_affordance} presents the evaluation results on the AGD20K Dataset~\cite{luo2022learning} using the \textit{Hard} split as defined by~\cite{qian2024affordancellm}. Remarkably, although our \texttt{ManipVQA} framework is trained solely on robotic affordances, such as grasping, it still demonstrates competitive performance on the general affordance grounding task. Notably, our method achieves the best NSS score, indicating strong model performance, but also the highest KLD, which is undesirable. We attribute this high KLD to the tendency of the SAM-series models to over-segment images. Additionally, the discrepancy between the ground truth represented as a heatmap, and our output, a segmentation mask, likely contributes to the lower KLD score. Our method also shows a promising ability to distinguish between fine-grained affordances associated with the same object class. For example, it can differentiate between \textit{Cut-with} and \textit{Hold} actions for a knife, which involve the blade and handle, respectively, as illustrated in Fig.~\ref{fig:affordance_knife_with_different_instr}.

\begin{figure*}[!htb]
\centering
\includegraphics[width=1\linewidth]{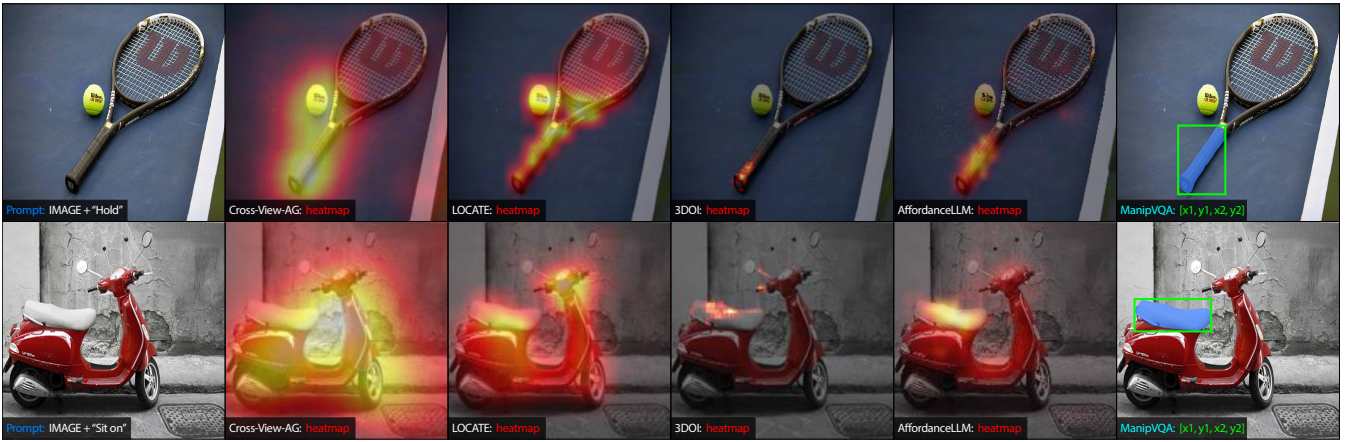}
\caption{Ilustration for the general affordance grounding tasks on the AGD20K~\cite{luo2022learning}. Notably, the action \textit{Sit-On} is unseen in our affordance training split. Blue masks are produced by SAM-HQ based on \texttt{ManipVQA}'s predicted bounding boxes. }
\vspace{-0.175cm}
\label{fig:general_affordance}
\end{figure*}

\begin{table}[]
\centering
\resizebox{0.48\textwidth}{!}{%
\begin{tabular}{c|c|c|c} \hline

\textbf{Method} &
\textbf{SIM $\uparrow$} &
\textbf{NSS $\uparrow$} &
\textbf{KLD $\downarrow$}  \\ \hline \hline

\multicolumn{1}{l|}{Cross-View-AG~\cite{luo2022learning}} & 
0.209 &
0.138 &
2.092 \\ 

\multicolumn{1}{l|}{LOCATE~\cite{li2023locate}} & 
0.282 &
0.276 &
1.829 \\

\multicolumn{1}{l|}{3DOI~\cite{qian2023understanding}} & 
0.200 &
0.549 &
4.017 \\ 

\multicolumn{1}{l|}{AffordanceLLM~\cite{qian2024affordancellm}} & 
\textbf{0.361} &
0.947 &
\textbf{1.661} \\ \hline

\multicolumn{1}{l|}{\textbf{Ours}} & 
0.246 &
\textbf{1.735} &
12.67 \\ \hline \hline

\end{tabular}%
}
\vspace{-0.2cm}
\caption{General affordance evaluation results on AGD20K~\cite{luo2022learning}.}
\label{tab:general_affordance}
\vspace{-0.6cm}
\end{table}

\subsubsection{\textbf{Robotic Manipulation in Simulator}}
Table~\ref{tab:simulator_evalution} illustrates the zero-shot performance of our model in SAPIEN~\cite{xiang2020sapien} when combined with a basic heuristic-based control policy. The model's success is largely due to the retention of commonsense reasoning capabilities in the MLLM and the incorporation of affordance knowledge, which enables effective robotic manipulation without prior fine-tuning on the task-specific data.

\begin{figure}[th]
  \centering
   \includegraphics[width=1\linewidth]{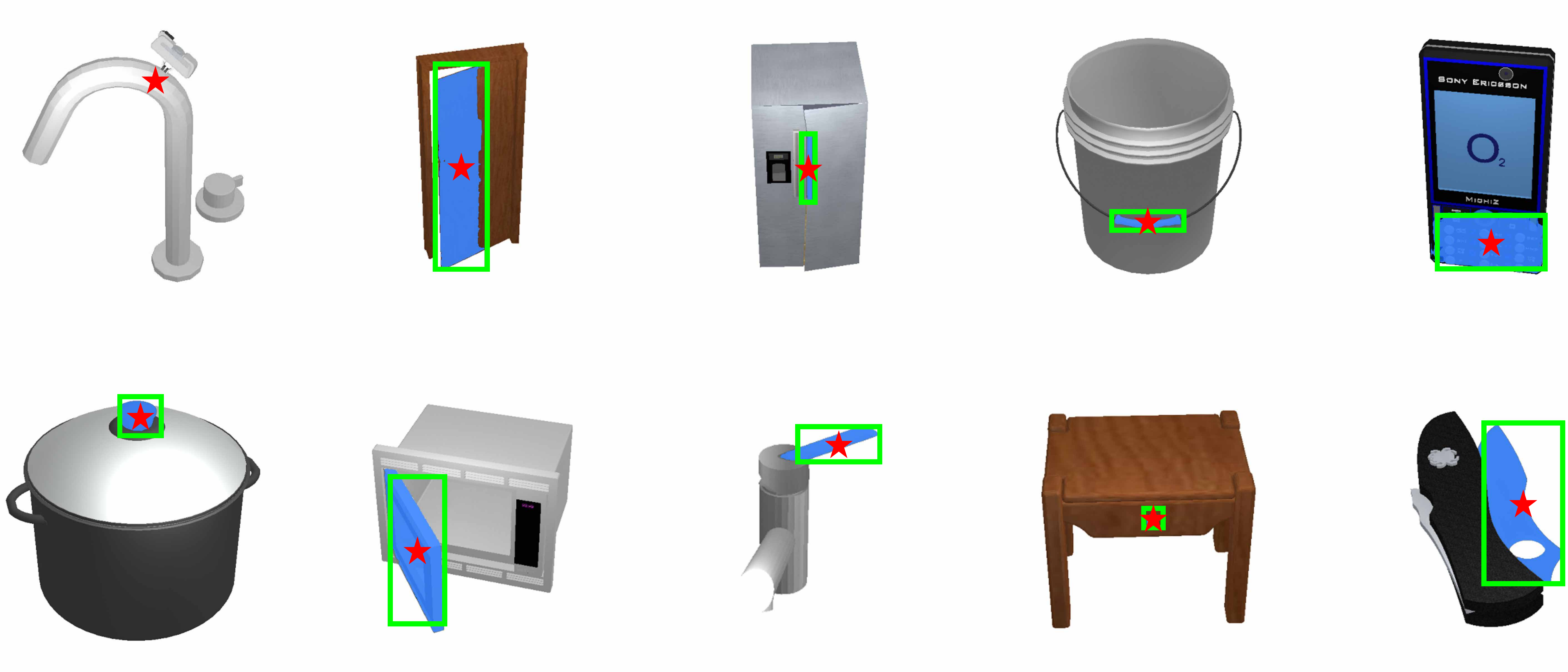}
   \caption{Visualizatios of \texttt{ManipVQA}'s predictions within the SAPIEN simulator. The first sub-image depicts the experimental setup, where the model predicts the gripper's contact point, represented by a red star, to achieve the desired movement of a specific object part. The green bounding boxes denote \texttt{ManipVQA}'s predictions, while the blue masks are obtained using SAM-HQ. The contact point is heuristically determined as the geometric center of the blue mask.}
   \label{fig:simulator_demos}
\vspace{-0.6cm}
\end{figure}

\begin{table*}[tb]
	\begin{center}
	\small
	\setlength{\tabcolsep}{1.5mm}{
	\begin{tabular}{c| cc c c c c c c c c c c c c c c}
    
    \hline
	\multirow{2}{*}{\textbf{}}&\multirow{2}{*}{\textbf{}} & \multicolumn{15}{c}{\textbf {Training Categories}}\\
 
 \textbf{Method}
    & \includegraphics[width=0.04\linewidth]{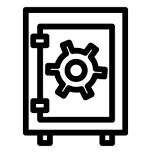}
    & \includegraphics[width=0.04\linewidth]{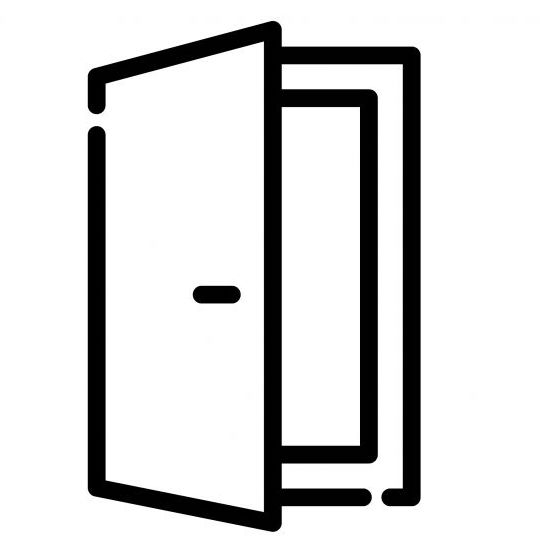}
    & \includegraphics[width=0.04\linewidth]{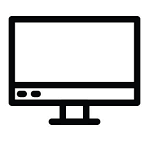}
    & \includegraphics[width=0.04\linewidth]{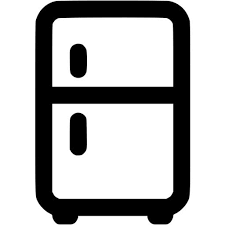}
    & \includegraphics[width=0.04\linewidth]{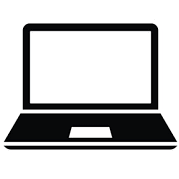}
    & \includegraphics[width=0.04\linewidth]{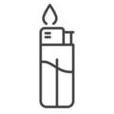}
    & \includegraphics[width=0.04\linewidth]{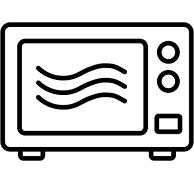}
    & \includegraphics[width=0.04\linewidth]{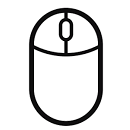}
    & \includegraphics[width=0.04\linewidth]{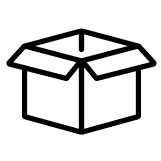}
    & \includegraphics[width=0.04\linewidth]{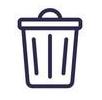}
    & \includegraphics[width=0.04\linewidth]{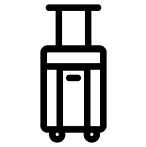}
    & \includegraphics[width=0.04\linewidth]{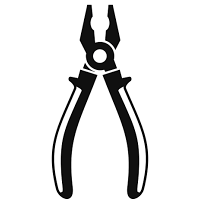}
    & \includegraphics[width=0.04\linewidth]{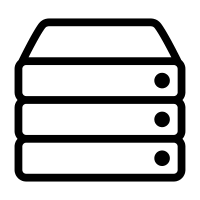}
    & \includegraphics[width=0.04\linewidth]{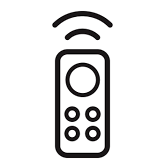}
    & \includegraphics[width=0.04\linewidth]{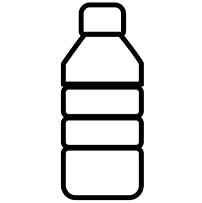}\\ \hline \hline

    \multicolumn{1}{l|}{Where2Act~\cite{mo2021where2act}} & 
    0.26 & 
    0.36 & 
    0.19 & 
    0.27 & 
    0.23 & 
    0.11 & 
    0.15 & 
    0.47 & 
    0.14 & 
    0.24 & 
    0.12 & 
    0.56 & 
    0.68 & 
    0.07 & 
    0.40 \\ 
  
    \multicolumn{1}{l|}{FlowBot3D~\cite{eisner2022flowbot3d}} & 
    0.67 & 
    0.55 & 
    0.20 & 
    0.32 & 
    0.27 & 
    0.31 & 
    0.61 & 
    0.68 & 
    0.15 & 
    0.28 & 
    0.18 & 
    0.21 & 
    0.70 & 
    0.18 & 
    0.26 \\
  
    \multicolumn{1}{l|}{ManipLLM~\cite{li2023manipllm}} & 
    \textbf{0.68} & 
    0.64 & 
    0.36 & 
    0.77 & 
    0.43 & 
    \textbf{0.62 } & 
    0.65 & 
    0.61 & 
    \textbf{0.65} & 
    0.52 & 
    0.40 & 
    0.64 & 
    \textbf{0.71} & 
    \textbf{0.60} & 
    \textbf{0.64} \\ \hline 
  

    \multicolumn{1}{l|}{\textbf{Ours}} & 
    0.67 & 
    \textbf{0.87} & 
    \textbf{0.46} & 
    \textbf{0.91} & 
    \textbf{0.56} & 
    0.42 & 
    \textbf{0.69} & 
    \textbf{0.79} & 
    0.41 & 
    0.53 & 
    \textbf{0.69} & 
    \textbf{1.00} & 
    0.53 & 
    0.17 & 
    0.58  \\ \hline \hline
	
   \hline
   \multirow{2}{*}{\textbf{}} & 
   \multirow{2}{*}{\textbf{}} & 
   \multicolumn{4}{c|}{\textbf {Training Categories}} &
   \multicolumn{9}{c|}{\textbf {Testing Categories}} \\
			
\textbf{Method}
    & \includegraphics[width=0.04\linewidth]{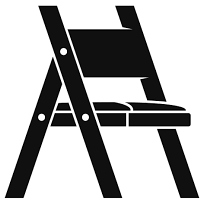}
    & \includegraphics[width=0.04\linewidth]{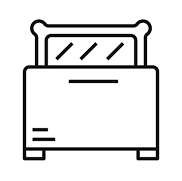}
    & \includegraphics[width=0.04\linewidth]{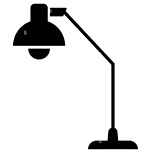}
    & \includegraphics[width=0.04\linewidth]{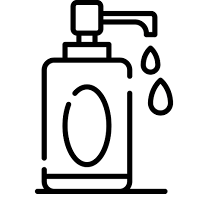}
    & \multicolumn{1}{c|}{\includegraphics[width=0.035\linewidth]{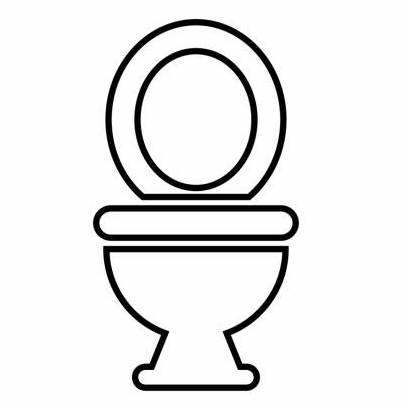}}
    & \includegraphics[width=0.04\linewidth]{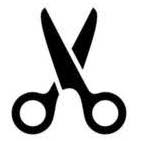}
    & \includegraphics[width=0.04\linewidth]{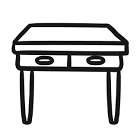}
    & \includegraphics[width=0.04\linewidth]{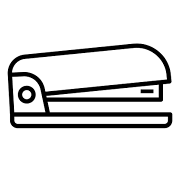}
    & \includegraphics[width=0.04\linewidth]{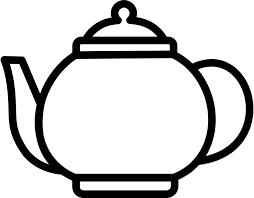}
    & \includegraphics[width=0.04\linewidth]{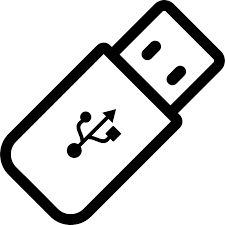}
    & \includegraphics[width=0.04\linewidth]{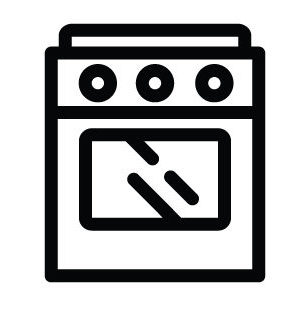}
    & \includegraphics[width=0.04\linewidth]{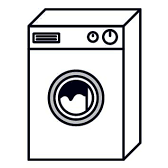}
    & \includegraphics[width=0.04\linewidth]{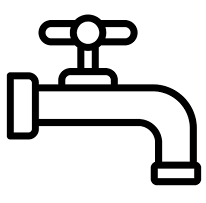}
    & \multicolumn{1}{c|}{\includegraphics[width=0.04\linewidth]{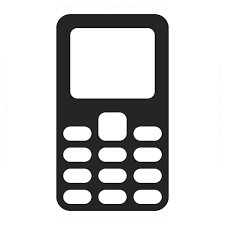}}
    & \multirow{-3}{*}{\textbf{AVG}} \\ \hline \hline
    
    \multicolumn{1}{l|}{Where2Act ~\cite{mo2021where2act}} & 
    0.13 & 
    0.18 & 
    0.13 & 
    0.40 & 
    \multicolumn{1}{c|}{0.18} & 
    0.35 & 
    0.38 & 
    0.28 & 
    0.05 & 
    0.21 & 
    0.17 & 
    0.20 & 
    0.15 & 
    0.15 & 
    \cellcolor[HTML]{C0C0C0} 0.25 \\ 
  
    \multicolumn{1}{l|}{FlowBot3D~\cite{eisner2022flowbot3d}} & 
    0.17 & 
    0.53 & 
    0.29 & 
    0.42 & 
    \multicolumn{1}{c|}{0.23} & 
    0.10 & 
    0.60 & 
    0.39 & 
    0.27 & 
    0.42 & 
    0.28 & 
    0.51 & 
    0.13 & 
    0.23 & 
    \cellcolor[HTML]{C0C0C0} 0.35 \\
  
    \multicolumn{1}{l|}{ManipLLM~\cite{li2023manipllm}} & 
    \textbf{0.41} & 
    \textbf{0.75} & 
    0.44 & 
    0.67 & 
    \multicolumn{1}{c|}{0.38} & 
    0.22 & 
    0.81 & 
    \textbf{0.86} & 
    \textbf{0.38} & 
    \textbf{0.85} & 
    0.42 & 
    \textbf{0.83} & 
    0.26 & 
    0.38 & 
    \cellcolor[HTML]{C0C0C0} 0.57 \\ \hline 
  
    \multicolumn{1}{l|}{\textbf{Ours}} & 
    0.20 & 
    0.56 & 
    \textbf{0.47} & 
    \textbf{0.75} & 
    \multicolumn{1}{c|}{\textbf{0.68}} & 
    \textbf{0.93} & 
    \textbf{0.92} & 
    0.82 & 
    0.32 & 
    0.58 & 
    \textbf{0.71} & 
    0.81 & 
    \textbf{0.69} & 
    \textbf{0.51} & 
    \cellcolor[HTML]{C0C0C0} \textbf{0.63} & \\ \hline \hline

	\end{tabular}}

    \caption{Performance evaluation within the SAPIEN simulator using PartNet-Mobility Dataset. Notably, while the baseline methods use distinct training and testing splits, our model achieves robust performance without fine-tuning on the SAPIEN samples.}
 
 \vspace{-0.675cm}
 \label{tab:simulator_evalution}	
 \end{center}
\end{table*}







\subsection{\textbf{Further Analysis}}

\subsubsection{\textbf{Ablation Studies}}
To dissect the contributions of the designs in \texttt{ManipVQA}, we performed a series of ablation studies. The results are presented in Table~\ref{tab:ablation_studies}. It was observed that the SOTA MLLM, SPHINX, cannot execute vision-based reasoning tasks in robotics without the \texttt{ManipVQA} Dataset. Furthermore, without the mixture of general vision data during the fine-tuning process, there is a noticeable decline in its physical understanding and affordance reasoning capabilities. This suggests that the current \texttt{ManipVQA} Dataset alone may not provide sufficiently large or diverse samples for effectively fine-tuning an MLLM.  Additionally, the absence of visual ensembles leads to a significant drop in the model's ability to reason about affordances, likely because robotic affordance reasoning often requires detailed part-level understanding.

\begin{table}[!hbpt]
\centering
\resizebox{0.48\textwidth}{!}{%
\begin{tabular}{c|c|c|c|c|c} \hline

\textbf{Manip. Data} &
\textbf{Vis. Ens} &
\textbf{Mix. Train} &
\textbf{Phys. } &
\textbf{Aff. Box} &
\textbf{Aff. Mask}  \\ \hline \hline

 & \checkmark & & 39.7 & 0.16 & 0.31 \\ 
\checkmark & \checkmark & 
  &
84.2 &
0.48 &
0.30 \\ 

 \checkmark & & 
\checkmark &
86.7 &
0.40 &
0.61 \\ 

\checkmark & 
\checkmark & 
\checkmark &
\textbf{90.3} &
\textbf{0.64} &
\textbf{0.62} \\ \hline \hline

\end{tabular}%
}

\caption{Ablation Studies. \textit{``Manip. Data"} indicates the use of the \texttt{ManipVQA} Dataset, \textit{``Vis. Ens"} represents the employment of a visual encoder ensemble, and \textit{``Mix. Train"} refers to the inclusion of a mixed general visual dataset during fine-tuning. \textit{``Phys."} assesses the model's physical grounding; and \textit{``Aff."} denotes the model's affordance reasoning capabilities.}
\label{tab:ablation_studies}
\vspace{-0.5cm}
\end{table}
\subsubsection{\textbf{Impact on Pre-Existing Vision Reasoning Ability}}
When fine-tuning a model on a specialized dataset like \texttt{ManipVQA}, it is essential to consider the potential impact on its pre-established general vision reasoning skills. To investigate this, we evaluated the model on the val split of RefCOCO+~\cite{kazemzadeh2014referitgame}. The post-fine-tuning accuracy was recorded at 81.8\%. This performance is to be compared against the accuracy of the pre-trained model from~\cite{gao2024sphinx} at 86.6\%. Despite a slight drop, the model retains a robust general vision reasoning ability.



\section{\textbf{Conclusion}}
\label{sec:conclusion}
This study seeks to reconcile the disparity between the capabilities of existing MLLMs and the demands of robotic systems. We present \texttt{ManipVQA}, a novel approach designed to equip MLLMs with manipulation-centric knowledge via a visual question-answering paradigm. Our approach involves the collection of a diverse set of images featuring interactive objects, thus encompassing a broad range of challenges related to object detection, affordance, and physical concept prediction. Empirical assessments performed in robotic simulators and across various vision task benchmarks substantiate the efficacy and resilience of \texttt{ManipVQA}.



\bibliographystyle{IEEEtran}
\bibliography{IEEEabrv,iros}

\end{document}